\documentclass[runningheads]{llncs}

\usepackage[T1]{fontenc}
\usepackage{graphicx}
\usepackage{amsmath}
\usepackage{amssymb}
\usepackage{booktabs}
\usepackage{caption}
\usepackage{subcaption}
\usepackage{multirow}
\usepackage{multicol}
\usepackage{hyperref}
\usepackage{xcolor}
\usepackage{comment}
\usepackage{cite}
\usepackage{letterspace}
\usepackage{fancyhdr}

\usepackage{pifont}
\definecolor{darkgreen}{HTML}{18864b}
\definecolor{darkred}{HTML}{bd451a}
\newcommand{\cmark}{\textcolor{darkgreen}{\ding{51}}}%
\newcommand{\xmark}{\textcolor{darkred}{\ding{55}}}%

\definecolor{faded}{HTML}{9e9e9e}
\colorlet{grey}{gray!60}

\renewcommand{\rightmark}{Rule-Based RL for Document Image Classification with VLMs}

\makeatletter
\AtBeginDocument{
  \hypersetup{
    pdftitle={\@title},
    pdfauthor={Michael Jungo and Andreas Fischer}
  }
}
\makeatother

\begin{document}

\title{Rule-Based Reinforcement Learning for Document Image Classification with Vision Language Models}
\titlerunning{Rule-Based RL for Document Image Classification with VLMs}

\author{%
Michael Jungo\inst{1, 2}%\orcidID{0009-0001-1790-1687}
\and
Andreas Fischer\inst{1, 2}%\orcidID{0000-0003-0069-3436}
}

\authorrunning{M. Jungo et al.}

\institute{%
  University of Applied Sciences and Arts Western Switzerland \\
  \email{\{michael.jungo,andreas.fischer\}@hefr.ch}
  \and
  University of Fribourg, Switzerland \\
  \email{\{michael.jungo,andreas.fischer\}@unifr.ch}
}

\maketitle

\begin{abstract}
  Rule-based reinforcement learning has been gaining popularity ever since DeepSeek-R1 has demonstrated its success
  through simple verifiable rewards. In the domain of document analysis, reinforcement learning is not as prevalent,
  even though many downstream tasks may benefit from the emerging properties of reinforcement learning, particularly
  the enhanced reason capabilities. We study the effects of rule-based reinforcement learning with the task of
  Document Image Classification which is one of the most commonly studied downstream tasks in document analysis. We
  find that reinforcement learning tends to have better generalisation capabilities to out-of-distritbution data,
  which we examine in three different scenarios, namely out-of-distribution images, unseen classes and different
  modalities. Our code is available at \href{https://github.com/jungomi/vision-finetune}{https://github.com/jungomi/vision-finetune}.

\end{abstract}

\thispagestyle{fancy}
\renewcommand{\headrulewidth}{0pt}
\lfoot{\small{\textit{Proceedings of the $5^{th}$ International Workshop on Machine Learning at the International Conference on Document Analysis and Recognition 2025 (ICDAR 2025)}}}

\section{Introduction}

Document Image Classification remains an important task as a first step for many document analysis approaches. It can be
as simple as deciding to which department it needs to be sent to, up to more complex systems that perform automated
analyses to extract and process the specific contents of various types of documents.

With the rise of popularity in Large Language Models (LLMs) and their steadily improving capabilities across multiple
modalities, they have become of interest for a wide variety of tasks. A key reason for this, is their ability of
in-context learning~\cite{few-shot-learner}, which makes them adaptable to any new task with only a few examples.
Many different Vision Language Models (VLMs)~\cite{vlm-survey-2024, vlm-survey-2025} are available, with commercial
models like GPT-4-Vision~\cite{gpt-4} and Gemini~\cite{gemini}, but also open-source models such as Llama
3.2-Vision~\cite{llama-3}, Qwen-2.5-VL~\cite{qwen2-5-vl} and Gemma 3~\cite{gemma-3}.
Even though the training data of most models is rarely fully disclosed, they are expected to also have been pre-trained
on document understanding, including publicly available datasets for document visual question answering~\cite{doc-vqa,
infographic-vqa, science-qa, ocr-vqa, visual-mrc}, as well as synthetic datasets~\cite{docmatix, pixmo, cosyn}.
These models are therefore equipped with a good understanding of documents and provide an excellent starting point for
Document Image Classification.

The RVL-CDIP~\cite{rvl-cdip} dataset is the most commonly used benchmark for Document Image Classification. It consist
of 400\,000 images of scanned documents across 16 classes, which is a labelled subset of the IIT-CDIP collection of
tobacco litigation documents~\cite{cdip-tobacco}. Over the years, many different models have been evaluated on this
dataset, but LLMs, including VLMs, have not been explored in much detail.

The contributions of this paper can be summarised as follows:

\begin{itemize}
  \item We show that reinforcement learning (RL) can be used as an alternative to supervised fine-tuning (SFT) for document image classification.
  \item We compare the generalisation capabilities of RL and SFT in three scenarios: out-of-distribution images, unseen
    classes and different modalities.
  \item We examine the effect of the reasoning ability that is induced by RL.
\end{itemize}

\section{Related Work}

\subsection{Rule-Based Reinforcement Learning for Vision Language Models}

After the success of rule-based reinforcement learning achieved by DeepSeek-R1~\cite{deepseek-r1}, naturally, the desire
of applying it to other models emerged and researchers started applying it to other domains and modalities.
R1-onevision~\cite{r1-onevision}, Vision-R1~\cite{vision-r1} and VLM-R1~\cite{vlm-r1} successfully applied the R1 style
reinforcement learning to vision language models shortly after the release of DeepSeek-R1 with the general consensus
that rule-based RL on its own can achieve competitive performance compared to SFT or even surpass it. Zhou et
al.~\cite{r1-zero-visual} managed to reproduce the ``aha-moment'' from DeepSeek-R1 in the form of visual reasoning
in a 2B VLM without any prior SFT. Jigsaw-R1~\cite{jigsaw-r1} applies RL to jigsaw puzzles, under the premise that the
reassembling of shuffled patches provide visual understanding that is transferable to downstream tasks, and found that
the model is able to generalise to other visual tasks that require spatial reasoning.

Rule-based RL has found its way to many other visual task, for example in the medical domain, Med-R1~\cite{med-r1} and
MedVLM-R1~\cite{medvlm-r1} concurrently investigated rule-based reinforcement learning, where they observed that it
improves the generalisation and reliability of the VLM across eight distinct medical imaging modalities, such as MRI, CT
or X-ray. To the best of our knowledge, downstream tasks involving document images have not yet been studied, which
compels us to explore it in the context of document image classification.

\subsection{Document Classification with Large Language Models}

Scius-Bertrand et al.~\cite{zero-shot-rvl-cdip} utilised LLMs to classify a subset of the RVl-CDIP with zero-shot and
one-shot prompting as well as fine-tuning them. At the time, VLMs were a novelty and only GPT-4-Vision was available,
hence they focused primarily on classifying the documents based on their textual content that was extracted with an OCR
engine.

The zero-shot and one-shot already achieved good performances, with the highest being GPT-4 at 61.8\% accuracy when only
given the OCR as input and 69.9\% when the images are included in the input. When they fine-tuned Mistral on the 1\,600
training samples from that subset, its accuracy increased from 45.4\% up to 83.4\%. They showed that fine-tuning with
a relatively small subset, compared to the full dataset, improves the results considerably.

Given the fact that the image based classification has achieved better results compared to the OCR, it would be expected
that fine-tuning a VLM has even more potential. We use this opportunity to study the fine-tuning of VLMs on the subset
they created, which ensures that we have a dataset that is sufficiently large to see the impact of the training while
not being too large to the point of not being able to finish the experiments in a reasonable time.

\section{Fine-Tuning Methods}

There are multiple stages in the pre-training of LLMs. After being trained on large unsupervised data, where the goal is
a simple next token prediction in order to learn fundamental text understanding and intricacies of the language, there
are two primary stages to shape the LLMs into the adaptable form that is commonly used for all sorts of tasks. Firstly,
supervised fine-tuning (SFT) is used to learn specific answers to a given question, this also includes instruction
following, which teaches the model to respond to a query rather than just completing the input by adding additional text
that is most likely to follow, as it was done during the initial pre-training. Afterward, reinforcement learning (RL)
is employed to steer the model's responses into a more preferential form, which may be stylistic choices, but also for
safety~\cite{safety-harmless-assistant, safety-llm-survey, safety-rlhf} by avoiding inappropriate phrases or by refusing
to answer unconscionable queries.

While those are well established practices, most downstream tasks, e.g. document classification, only apply additional
SFT to adapt the model to their specific tasks. One of the most prevalent reasons is the fact that RL requires a reward
and value model to be trained alongside the LLM itself. This not only makes it much more demanding in terms of resources,
but also adds a training complexity that made it unsuitable for downstream fine-tuning. With the release of
DeepSeek-R1~\cite{deepseek-r1}, reinforcement learning with verifiable rewards (RLVR)~\cite{rlvr-tulu-3, rlvr-incentive,
rlvr-visual-rft} has gained a lot of attention, as they have shown that their Group Relative Policy Optimisation
(GRPO)~\cite{deepseek-math} training method can achieve excellent results by replacing the reward and value models with
simple verifiable reward functions. For downstream tasks such as classification, it is straightforward to define
a reward function, making it much more accessible and a viable option.

\subsection{Supervised Fine-Tuning (SFT)}

Supervised fine-tuning has been the de facto standard fine-tuning method for classical tasks with a clear cut answer,
such as classification, where the answer is restricted to one of the known classes. In the context of LLMs, SFT not only
serves to adapt the model to the given task but also to enforce an expected output format, which makes it easy to parse
the expected class. In the simplest form, the model should reply purely with the predicted class without any additional
explanation. While that can be achieved through more restrictive prompts, the fine-tuning bakes it into the model,
making it more reliable. This is achieved with a simple cross-entropy loss, which is applied to the tokens of the
response in order to alternate the model's parameters such that the likelihood of producing the desired class is
increased. The loss is applied exclusively to the response, meaning that only the tokens in the response are learned.

\subsection{Reinforcement Learning (RL)}

While reinforcement learning has been included in the training of LLMs, it is primarily used to align the responses
with human preferences, where annotators are presented with multiple responses which they have to rank in order of
preference. As having a human in the loop would be prohibitively expensive, reinforcement learning from human feedback
(RLHF)~\cite{rlhf} trains a reward model that estimates the human preferences based on the collected preference
annotations for a dataset with curated responses. Needing an additional reward model, as well as a dataset with
annotated preferences, makes it unappealing for most downstream tasks, particularly in a low-resource scenario.

\subsubsection{Group Relative Policy Optimisation (GRPO).}

Group Relative Policy Optimisation (GRPO) removed the reward model entirely and replaced it with simple verifiable
rewards, which can be implemented with any deterministic function that can verify the quality of a response and assign
it a reward value. This opens up a lot of possibilities for downstream tasks with a verifiable outcome, for example, the
classification can easily be verified by checking whether the output corresponds to the expected class.

To get the same effect as the human preference ranking, for each query $q$ a group of responses $\{o_{1}, \ldots,
o_{G}\}$ are sampled from the old policy $\pi_{\theta_{old}}$ and the relative advantages within the group $G$ of
responses is calculated. GRPO optimises the policy $\pi_{\theta}$ to maximise the following objective:

\begin{equation}\label{eq:grpo}
  \scalebox{0.84}{$
  \begin{aligned}
    \mathcal{J}_{GRPO}(\theta) &= \mathbb{E}_{q \sim P(Q)},\{o_{i}\}_{i = 1}^{G} \sim \pi_{\theta_{old}}(\cdot \mid q) \\
      &\Biggl[ \frac{1}{G}\sum_{i=1}^{G}{\frac{1}{\lvert o_{i} \rvert}\sum_{t=1}^{\lvert o_{i} \rvert}{\biggl\{min \left( \frac{\pi_{\theta}(o_{i} \mid q)}{\pi_{\theta_{old}}(o_{i} \mid q)} A_{i}, clip \left(  \frac{\pi_{\theta}(o_{i} \mid q)}{\pi_{\theta_{old}}(o_{i} \mid q)}, 1 - \epsilon, 1 + \epsilon \right) A_{i} \right)}} \\
      &-\beta \mathbb{D}_{KL}\left( \pi_{\theta}(\cdot \mid q) \| \pi_{ref}(\cdot \mid q) \right) \biggr\} \Biggr]
  \end{aligned}
  $}
\end{equation}

\begin{equation}\label{eq:grpo-advantage}
A_{i} = \frac{r_{i} - mean(\{r_{1}, r_{2}, \ldots, r_{G}\})}{std(\{r_{1}, r_{2}, \ldots, r_{G}\})}
\end{equation}

where $A_{i}$  is the group-normalised advantage of the sample $i$, which is calculated by comparing the reward $r_{i}$
to the entire group. Concretely, this means that a response that is better than the average response from that group
will produce a positive advantage, which encourages the model to choose this particular response, as the objective
becomes larger. For the opposite case, when the response is worse than the average response of the group, it will result
in a negative advantage, which discourages the model by reducing the likelihood of producing that response.
The $\mathbb{D}_{KL}$ term refers to the KL divergence between the distribution of the current policy $\pi_{\theta}$ and the
reference policy $\pi_{ref}$, which ensures that the model does not deviate too far from the original model. $\beta$ and
$\epsilon$ are hyperparamters, which are the coefficient for the KL penalty and the clipping threshold, respectively.

\subsubsection{Reinforcement Learning with Verifiable Rewards (RLVR).} 

In RLVR the rewards are calculated with rule-based reward functions. These are for the most part simple checks, which
evaluate the quality of the response. In the case of classification, it can be a binary check for whether it is correct,
in which case the reward would be 1.0 if it is correct and otherwise 0.0. As the reward is any real valued number,
a reward can span an entire spectrum, where negative values would be a penalty. An example for continuous values would
be length reward, where each additional token increases the reward. If such a reward function is chosen, the magnitude
of the value needs to be managed, particularly to not overpower other rewards which might be fixed.

We chose two reward functions, namely a format reward to ensure that the final classification can be easily extracted,
while also enforcing the inclusion of a reasoning trace, and one for the classification accuracy, which is a simple
check of whether the predicted class was correct.

\begin{itemize}
  \item \textbf{Format}: To ensure that the predicted class can be easily extracted from the response, the class must
    be given inside an \texttt{<answer></answer>} tag. To make the format a little more strict and promote providing
    the reasoning steps for the decision, the response also needs to include a \texttt{<reasoning></reasoning>} tag.
    Each of the tags being present gives a reward of 0.5 while an additional 0.5 is awarded if they are given in the
    exact order of reasoning followed by answer. Any superfluous occurrence of the tags will induce a reward
    penalty, in the form of a negative penalty of $-0.5$ for every additional occurrence.
  \item \textbf{Classification}: Once the predicted class has been successfully extracted from the response, a reward
    of 1.0 is given if the classification was correct, otherwise it is 0.0.
\end{itemize}

\subsection{Parameter Efficient Fine-Tuning (PEFT)}

Due to the large number of parameters in LLMs, full fine-tuning is prohibitively costly as it demands a lot of GPU
resources. To address this issue, parameter efficient fine-tuning (PEFT) methods have been introduced, where only
a small number of extra parameters are fine-tuned while keeping the original model's parameters untouched. Since only
the newly added parameters are fine-tuned, it requires a lot less memory for the gradients, as well as reducing the
storage requirements, as only these parameters need to be stored, rather than having to store the whole model again due
to minor adjustments of the base parameters.

One of the most used method is Low-Rank Adaptation (LoRA)~\cite{lora}, which injects a set of low-rank weight matrices,
called adapters, to already existing weights. For any adapted weight, the input is passed through the original weight as
well as the adapter, whose outputs are then combined by taking their sum. Instead of updating the original weight
in-place, the weight update is entirely reflected by the adapter. As the weight updates supposedly have a low intrinsic
rank~\cite{intrinsic-dimensionality, intrinsic-dimensionality-lm}, the adapters are decomposed into two low-rank matrices to further reduce the number of added parameters. Formally,
given a weight matrix $W_{0} \in \mathbb{R}^{d \times k}$, the weight update $\Delta W$ is decomposed into $BA$, where
$B \in \mathbb{R}^{d \times r}, A \in \mathbb{R}^{r \times k}$  with the rank $r \ll min(d, k)$. The output of this
layer is calculated as follows:

\begin{equation}\label{eq:lora}
  h(x) = W_{0}x + \Delta Wx = W_{0}x + BAx
\end{equation}

Even though this adds a slight latency during the training, the overall benefits are much greater and the latency
becomes negligible. Furthermore, the latency can be removed entirely for inference once the model has finished training,
by merging the adapters into the weight, due to the mathematical equivalence of $W_{0}x + BAx = (W_{0} + BA)x$.

The memory requirements for training can be additionally reduced by quantising the pre-trained model to 4-bit.
QLoRA~\cite{qlora} showed that weight quantisation to 4-bit precision did not sacrifice performance compared to the full
16-bit precision when LoRA adapters are fine-tuned, but substantially reduces the memory footprint of the model. We
therefore decided to use QLoRA, as it established itself as the commonly preferred PEFT method.

\section{Experiments}

To investigate the difference between SFT and RL and their ability to generalise to out-of-distribution data, we
consider three scenarios, that represent different types of previously unseen data. The first, more commonly studied
scenario is the evaluation on a dataset with the same classes but with images from a completely different source.
Whereas for the second scenario, we remove a fraction of the available document classes for the training and examine
whether the model can classify previously unseen classes. As LLMs are rather flexible in what classes they should
predict, due to the variable prompts, it raises the question whether the fine-tuning helps or hinders the adaptation to
unseen classes, considering the same task. The third scenario focuses on different modalities, specifically images and
text, where the same documents are given to the model either as an image or as their textual content, that was extracted
through an OCR system. Lastly, the reasoning traces that have been enabled by the RL and their effect on the
classification are examined.

\subsection{Experimental Setup}

All our experiments are based on
LLama-3.2-11B-Vision-Instruct\footnote{\href{https://huggingface.co/meta-llama/Llama-3.2-11B-Vision-Instruct}{https://huggingface.co/meta-llama/Llama-3.2-11B-Vision-Instruct}}
as the model that is fine-tuned using QLoRA~\cite{qlora}, either with SFT or RL. For both fine-tuning methods the LoRA
adapters are exclusively added to the LLM, therefore the vision encoder is untouched. We use a group size of $G = 8$ for
the RL, meaning that 8 responses are sampled for every input and the advantages are calculated based on these generated
samples. The KL divergence $\mathbb{D}_{KL}$ in \autoref{eq:grpo} requires the reference model $\pi_{ref}$, which is the
base model before fine-tuning. Since the base model is unaltered due to the use of LoRA adapters, we can access it
by temporarily disabling the LoRA adapters, which does not incur any additional memory requirement to store the
reference model.

\subsection{Out-of-Distribution Images}

The images from RVL-CDIP have a very particular look, which is typical for documents from the late
20\textsuperscript{th} century. Additionally, since they have been scanned at the time, it resulted in fairly low
quality images in today's standards, approximately 100dpi, as well as introducing noise and other artefacts.

In contrast, Larson et al.~\cite{rvl-cdip-ood} created a dataset containing more modern documents, where a large
proportion are born-digital instead of being scanned versions of physical documents. They followed the same annotation
strategy as RVL-CDIP with the same 16 classes and named it RVL-CDIP-N, where the \textit{N} stands for \textit{new
distribution}. The dataset contains 1\,002 images of documents that were collected from either web searches or the
public DocumentCloud\footnote{\href{https://documentcloud.org/}{https://documentcloud.org/}} repository.

It is worth noting that since these documents are publicly available on the internet, there is a chance some of them may
also have been included in the pre-training data of available models. However, it is much less likely that the
classification task was part of the pre-training. The same could also be said for the original RVl-CDIP dataset, but
born-digital documents are more likely to end up in the large pre-training datasets, as the PDF files provide readily
accessible information about the contents and structure of the documents.

\begin{table}[ht]
  \setlength{\tabcolsep}{10pt}
  \begin{center}
    \caption{
      \textbf{Out-of-Distribution Images.}
      LLama-3.2-11B-Vision-Instruct trained on the 1\,600 images of the original RVL-CDIP training set with either
      SFT or RL and evaluated the classification accuracy on the
      RVL-CDIP test set (in-distribution) and the RVL-CDIP-N test set (out-of-distribution).
    }\label{tab:results-rvl-cdip-N}
    \begin{tabular}[c]{c|cc}
      \toprule
      & RVL-CDIP & RVL-CDIP-N \\
      Training Method & \textit{ID} & \textit{OOD} \\
      \midrule
      SFT & 85.62 & 94.23 \\
      RL & 77.50 & 96.21 \\
      \bottomrule
    \end{tabular}
  \end{center}
\end{table}

We trained the model on the RVL-CDIP training set, once with a supervised fine-tuning (SFT) and once with reinforcement
learning (RL), and evaluated it on the accompanying RVL-CDIP test set, which contains in-distribution images, as well as
on out-of-distribution images from the RVl-CDIP-N test set. The results are shown in \autoref{tab:results-rvl-cdip-N}.

While the model trained with SFT has a better accuracy on the in-distribution images, the model trained with RL narrowly
surpasses it on the out-of-distribution images. Even though the difference between the two models is quite small in
absolute terms, the relative differences going from in-distribution to out-of-distribution is considerably larger in
favour of RL. This indicates that RL generalises better to new images that stray away from the characteristics found in
the training data.

The lower accuracy of the in-distribution result from RL compared to SFT might be due to some training instability, see
\autoref{sec:rl-training-instability} for the details about the encountered training instabilities.

\subsection{Unseen Classes}\label{sec:unseen-classes}

In order to evaluate the generalisation capabilities of the model to previously unseen classes, we withhold some classes
from the training. This is achieved by splitting the dataset into two subsets, where only 10 out of the 16 classes in
RVl-CDIP are kept for the training by excluding the remaining 6 classes. The test set follows the split of
classes listed in \autoref{tab:classes-split}, which allows evaluating the models on unseen classes, while still
keeping the images from the same distribution. Since the dataset has an equal number of images for every class, the
exact choice of classes has no effect on the size of the two subsets and were therefore randomly chosen.

\begin{table}[ht]
  \setlength{\tabcolsep}{10pt}
  \begin{center}
    \caption{
      \textbf{Classes Split.}
      The 16 classes of the RVL-CDIP dataset split into two subsets, the 10 in-distribution classes used for
      training, and the 6 remaining out-of-distribution classes, that are exclusively used for the evaluation.
    }\label{tab:classes-split}
    \begin{tabular}[c]{ll}
      \toprule
      \multicolumn{1}{c}{\textbf{10 classes}} & \multicolumn{1}{c}{\textbf{6 classes}} \\
      \multicolumn{1}{c}{\textit{ID}} & \multicolumn{1}{c}{\textit{OOD}} \\
      \midrule
      letter                 & email          \\
      form                   & handwritten    \\
      advertisement          & news article   \\
      scientific report      & invoice        \\
      scientific publication & presentation   \\
      specification          & questionnaire  \\
      file folder            &                \\
      budget                 &                \\
      resume                 &                \\
      memo                   &                \\
      \bottomrule
    \end{tabular}
  \end{center}
\end{table}

During the training, only the images from the 10 classes are shown to the model, while also only asking for these 
classes in the prompt. As a consequence, the model also needs to be able to adapt to a new prompt with new, previously
unseen, classes, rather than always answering with the classes it may have memorised during the training. To assess the
adaptability of the models, including their instruction following capabilities, every test set is evaluated with three
prompts, which contain all 16 classes, only the 10 classes used during training or the 6 unseen classes, respectively.
This means, that some combinations of prompts and test sets have no overlap in the classes, therefore if the model is
following the instructions correctly, it should always predict one of the provided classes. Specifically, when asked for
the 6 unseen classes in the prompt, but evaluated on the test set of the 10 classes it was trained on, the model should
not predict one of the classes it might have memorised during training.

\begin{table}[ht]
  \setlength{\tabcolsep}{5pt}
  \begin{center}
    \caption{
      \textbf{Unseen Classes.}
      LLama-3.2-11B-Vision-Instruct trained on the 1\,000 images of the 10 selected classes from the original
      RVL-CDIP training set, where the prompt contains only the 10 available classes.
      Each test set is evaluated with three variations of the prompt containing either all classes, the 10 classes
      seen during training or the 6 unseen classes, respectively.
      Values in \textcolor{grey}{grey} indicate that the classes in the prompt differ from the actual classes in the
      test data.\\
      ${}^\dagger$ Models that were trained on all classes as a reference.
    }\label{tab:results-unseen-classes}
    \begin{tabular}[c]{cr|ccc}
      \toprule
      & & \multicolumn{3}{c}{Test Data} \\
      Training Method & \multicolumn{1}{c|}{Prompt} & \textit{10 classes} & \textit{6 classes} & \textit{All classes} \\
      \midrule
      SFT on all classes${}^\dagger$ & All classes & \textcolor{grey}{87.50} & \textcolor{grey}{80.73} & 85.62 \\
      RL on all classes${}^\dagger$ & All classes & \textcolor{grey}{80.36} & \textcolor{grey}{74.48} & 77.50 \\
      \hline
      \multirow{3}{*}{SFT} & 10 classes & 89.29 &  \textcolor{grey}{0.00} & \textcolor{grey}{56.25} \\
        &   6 classes & \textcolor{grey}{76.79} & 43.23 & \textcolor{grey}{61.87} \\
        & All classes & \textcolor{grey}{89.29} & \textcolor{grey}{18.23} & 63.13 \\
      \hline
      \multirow{3}{*}{RL} &  10 classes & 90.18 &  \textcolor{grey}{0.00} & \textcolor{grey}{55.00} \\
       &   6 classes &  \textcolor{grey}{9.82} & 78.65 & \textcolor{grey}{32.50} \\
       & All classes & \textcolor{grey}{88.40} & \textcolor{grey}{58.33} & 78.75 \\
      \hline
      \multirow{3}{*}{RL after SFT} & 10 classes & 88.39 & \textcolor{grey}{0.00} & \textcolor{grey}{55.00} \\
       &   6 classes & \textcolor{grey}{54.46} & 48.44 & \textcolor{grey}{53.75} \\
       & All classes & \textcolor{grey}{91.07} & \textcolor{grey}{33.85} & 66.87 \\
      \bottomrule
    \end{tabular}
  \end{center}
\end{table}

The results in \autoref{tab:results-unseen-classes} show the different combinations of prompts and test sets for the 10
in-distribution classes, the 6 out-of-distribution classes and also all classes combined. The models trained with SFT
and RL are very close on the test set for the 10 classes they were trained on, indicating that both methods can achieve
equally good results. However, they start to deviate quite strongly when looking at the out-of-distribution classes,
where the SFT model drops to 43.23\% accuracy, i.e. less than half of the in-distribution accuracy, when evaluated on
the 6 unseen classes, compared to the 78.65\% of the RL model. A slight drop is always expected when moving to
previously unseen data, but the RL certainly managed to maintain much more of the performance than the SFT.

Another interesting aspect is the instruction following capabilities of the models. When the models are evaluated on the
test set with the 10 classes they were trained, but the prompt asks them to only choose from one of the 6 classes, they
should in theory achieve an accuracy of 0\%, since there is no overlap in classes between the prompt and the actual
data. Unfortunately, this is not the case, and they still occasionally respond with one of the classes they were trained
on, ignoring the ones provided in the prompt. While the RL is only doing it roughly 1 out of 10 cases, which achieves an
accuracy of 9.82\%, the SFT does it much more frequently, resulting in an accuracy of 76.79\%. The unexpected accuracy
points to the issue of the model having memorised the classes it was trained, and in fact, the SFT model only responds
with one of the asked classes in less than 2\% of the cases. The expected results would be 0\%, which can be clearly
observed in the inverse case, where the model is evaluated on the test set with the 6 unseen classes, but the prompt
contains only the 10 classes it was trained on, granted that the new classes were never presented to the model at any
point. This exposes a broader problem of SFT overfitting on the training data, which leads to memorisation instead of
generalisation, which is inline with the findings of Chu et al.~\cite{sft-vs-rl} but for document image classification.

Since it is common practice to apply SFT first and then RL afterwards, we also examined the effects of following this
strategy in the context of downstream tasks. RL after SFT alleviates the issue of memorisation to a certain degree,
where the previous 76.79\% accuracy of the mismatched classes is reduced to 54.46\%. There is a caveat to this change,
as the model responds in only roughly double the number of cases with the asked classes compared to before, but instead,
starts mangling classes, such as \textit{scientific journal article}, or completely deviating from any of the available
classes, e.g.\ \textit{appendix} or \textit{membership investment notice}. This means that the damaged caused by SFT is
not as easily reversible.

\subsection{New Modality}

Another form of generalisation is the change of modality. All models have been trained to classify document images,
however the same task could also be performed with the extracted text of the documents. Therefore, a model that has
learned the fundamental task of document classification, should also be able to perform that task on a different
modality to a certain extent, granted that the base model supports other modalities. In the case of Llama-3.2 Vision, it
is and extension of Llama-3.2 by adding a vision encoder that projects the images into embedding space of the base LLM,
therefore the underlying text understanding is still present, making it a great option to evaluate the two different
modalities.

\begin{table}[ht]
  \setlength{\tabcolsep}{10pt}
  \begin{center}
    \caption{
      \textbf{New Modality.}
      LLama-3.2-11B-Vision-Instruct trained on the 1\,600 images or their text contents, extracted through an OCR model,
      of the original RVL-CDIP training set.
      The classification accuracy is evaluated for each combination of modalities.
      Values in \textcolor{grey}{grey} indicate the test data having a different modality than the model was trained on.
    }\label{tab:results-modality}
    \begin{tabular}[c]{cc|cc}
      \toprule
      \multicolumn{2}{c|}{Training} & \multicolumn{2}{c}{Test Data} \\
      Method & Data & Image & OCR \\
      \midrule
      \multirow{2}{*}{SFT} & Image & 85.62 & \textcolor{grey}{60.62} \\
        & OCR & \textcolor{grey}{43.75} & 81.88 \\                     
      \hline
      \multirow{2}{*}{RL} & Image & 77.50 & \textcolor{grey}{52.50}  \\
       & OCR & \textcolor{grey}{25.00} & 71.25 \\                      
      \bottomrule
    \end{tabular}
  \end{center}
\end{table}

Contrary to the previous results, this time the model trained with SFT generalises better to the other modality than the
one with RL, as is evident by \autoref{tab:results-modality}. In either case, the drop-off is more severe when going
from OCR to images. The very steep drop-off from 71.25\% to 25\% of the RL model can be partially explained by the
training instabilities that are discussed in \autoref{sec:rl-training-instability}, because the responses contain
formatting issues such as missing the closing tag of the answer or not having any tags at all. The same cannot be said
for the other direction, going from images to OCR, which has only a few rare cases where the format was not respected,
hence it cannot be the sole reason. 

\subsection{Reasoning}

Increased test-time compute has been shown to improve the outputs of LLMs~\cite{scale-test-time-compute}.
Chain-of-Thought (CoT)~\cite{cot} is a technique to include a series of reasoning steps into the desired output of the
LLMs, which is intended to help the model get to the solution by breaking it down into smaller intermediate steps. Using
CoT in a supervised setting would require curating a dataset of high quality reasoning traces for the given task, which
is time and labour intensive. On the other hand, GRPO can integrate reasoning traces into the training by including them
in the generations of the samples and only enforcing the format, rather than the exact reasons the model is supposed to
give. There is no guarantee that the reasons are always correct, even if the answers are correct, but it still allows
the model to generate extra tokens for increased test-time compute (colloquially referred to as ``thinking time'').

To judge the impact of the reasoning at test-time, we compare the model trained with GRPO with and without the reasoning
in the answer. Inspired by Qwen3's~\cite{qwen3} ``non-thinking'' mode, we use the same model that was trained to include
reasoning, but prefilling the output with an empty reasoning tag, i.e. \texttt{<reasoning></reasoning>}, which forces
the model to go straight to the answer, as the reasoning was already completed.

\begin{table}[t]
  \setlength{\tabcolsep}{5pt}
  \begin{center}
    \caption{
      \textbf{Impact of Reasoning.}
      The same models that were trained on the 10 classes as in \autoref{tab:results-unseen-classes}, but
      comparing the models after RL with reasoning (\cmark) and without (\xmark), where the response is prefilled
      with an empty reasoning tag, i.e. \texttt{<reasoning></reasoning>}. Values in \textcolor{grey}{grey} indicate that
      the classes in the prompt differ from the actual classes in the test data.\\
    }\label{tab:results-reasoning}
    \begin{tabular}[c]{ccr|ccc}
      \toprule
      & & & \multicolumn{3}{c}{Test Data} \\
      Training Method & Reasoning & \multicolumn{1}{c|}{Prompt} & \textit{10 classes} & \textit{6 classes} & \textit{All classes} \\
      \midrule
      \multirow{6}{*}{RL} & \cmark &  10 classes & 90.18 &  \textcolor{grey}{0.00} & \textcolor{grey}{55.00} \\
       & \cmark &   6 classes &  \textcolor{grey}{9.82} & 78.65 & \textcolor{grey}{32.50} \\
       & \cmark & All classes & \textcolor{grey}{88.40} & \textcolor{grey}{58.33} & 78.75 \\
      \cline{2-6}
       & \xmark &  10 classes & 75.89 &  \textcolor{grey}{0.00} & \textcolor{grey}{46.25} \\
       & \xmark &   6 classes &  \textcolor{grey}{0.02} & 58.33 & \textcolor{grey}{23.75} \\
       & \xmark & All classes & \textcolor{grey}{71.43} & \textcolor{grey}{48.44} & 67.50 \\
      \hline
      \multirow{6}{*}{RL after SFT} & \cmark & 10 classes & 88.39 & \textcolor{grey}{0.00} & \textcolor{grey}{55.00} \\
       & \cmark &   6 classes & \textcolor{grey}{54.46} & 48.44 & \textcolor{grey}{53.75} \\
       & \cmark & All classes & \textcolor{grey}{91.07} & \textcolor{grey}{33.85} & 66.87 \\
      \cline{2-6}
       & \xmark & 10 classes & 91.07 & \textcolor{grey}{0.00} & \textcolor{grey}{55.00} \\
       & \xmark &   6 classes & \textcolor{grey}{59.82} & 50.00 & \textcolor{grey}{53.13} \\
       & \xmark & All classes & \textcolor{grey}{90.20} & \textcolor{grey}{32.30} & 66.25 \\
      \bottomrule
    \end{tabular}
  \end{center}
\end{table}

In \autoref{tab:results-reasoning} we can see that the accuracy of the model trained with pure RL deteriorates across
the board when the reasoning is disabled. This would suggest that the increased test-time compute is indeed helpful for
the model to make the correct classification, but we observed that the model starts to disregard the imposed format and
just puts everything inside the answer tag. This makes it impossible to extract the final classification in a reliable
manner and therefore undermines the fine-tuning efforts. The model is still able to adhere to the format and responds
with a singular answer tag containing only the predicted class, which may not necessarily always be the correct class,
but in roughly a third of the cases, that is no longer the case. In the most extreme scenario, namely when the model is
evaluated on the 10 classes but the prompt contains the other 6 classes, over half of the responses completely ignore
the format. It seems that the model is more likely to forego the format in scenarios where it struggles to find the
correct solution.

This pattern cannot be observed for the model that was first trained with SFT and then followed by RL. Removing the
reasoning has very little effect in comparison to the pure RL and more of often than not it is slightly improving the
accuracy. As we observed in \autoref{sec:unseen-classes}, the underlying behaviour of this model is predominantly
attributed to SFT, therefore the reasoning is an after-effect and removing it has no discernible impact, which arguably
brings it closer to the original SFT output structure.

\begin{figure}[t]
  \begin{center}
    \scriptsize
    \begin{tabular}[c]{p{68px}p{100px}p{68px}p{100px}}
      \begin{subfigure}[h]{68px}\centering \includegraphics[width=68px]{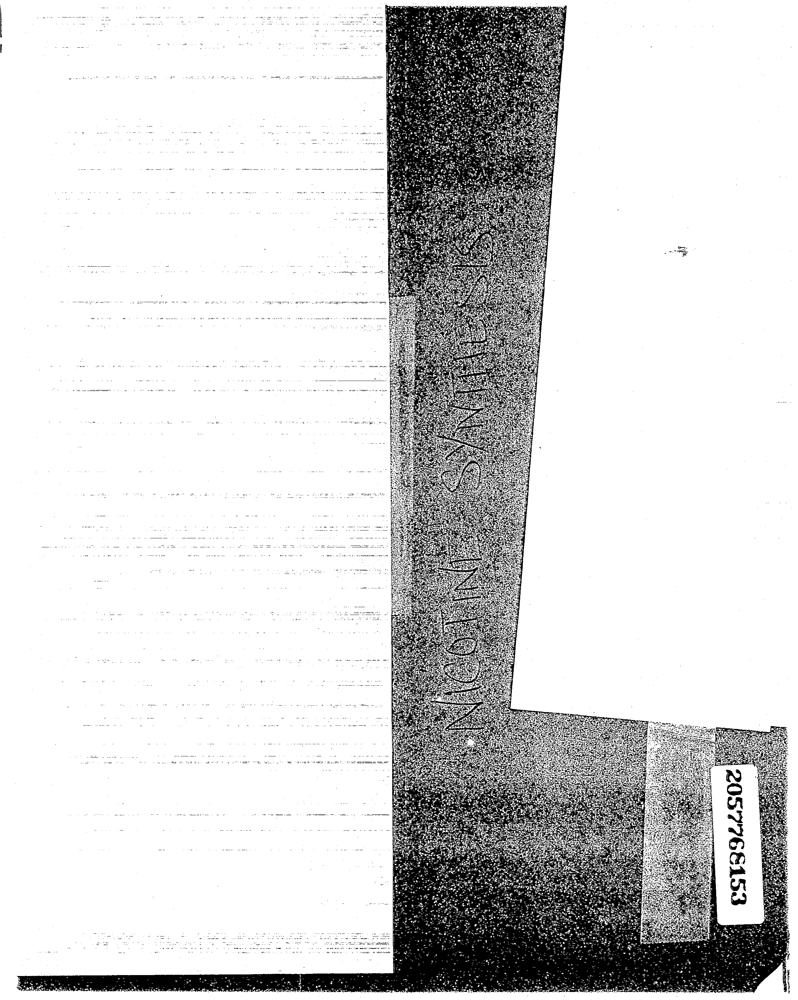} \\ GT: \hspace{0.5em}file folder\\Pred: \textcolor{darkgreen}{file folder}\captionsetup{font=scriptsize}\caption{}\end{subfigure} & \begin{minipage}{100px}\vspace{-65px}\textls[-25]{The image appears to be a file folder, as indicated by the rectangular shape with a tab on the bottom right corner and the presence of a label with a number, which is a common feature of file folders.}\end{minipage} &
      \begin{subfigure}[h]{68px}\centering\includegraphics[width=68px]{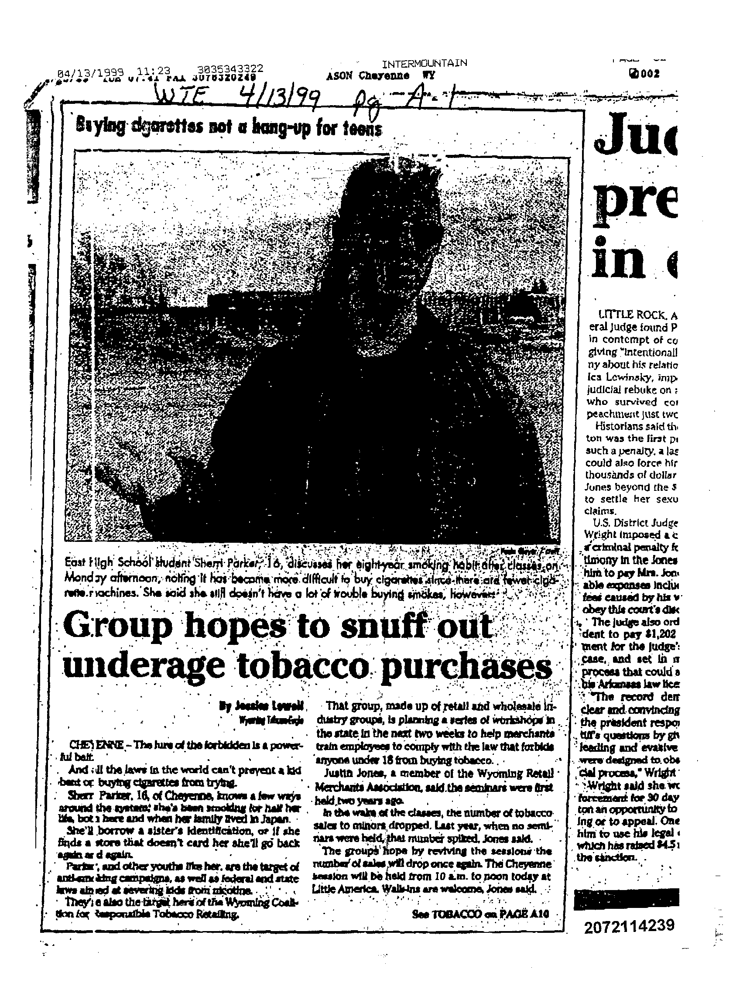}\\ GT: \hspace{0.5em}news article\\Pred: \textcolor{darkgreen}{news article}\captionsetup{font=scriptsize}\caption{}\end{subfigure} & \begin{minipage}{100px}\vspace{-25px}\textls[-25]{The image depicts a news article, which is a type of document that reports on current events or news stories. The content of the article includes a headline, text, and a photo of a person, which are all typical features of a news article. The layout and design of the page, including the use of columns and headlines, further support this classification.}\end{minipage} \\
      \begin{subfigure}[h]{68px}\centering\includegraphics[width=68px]{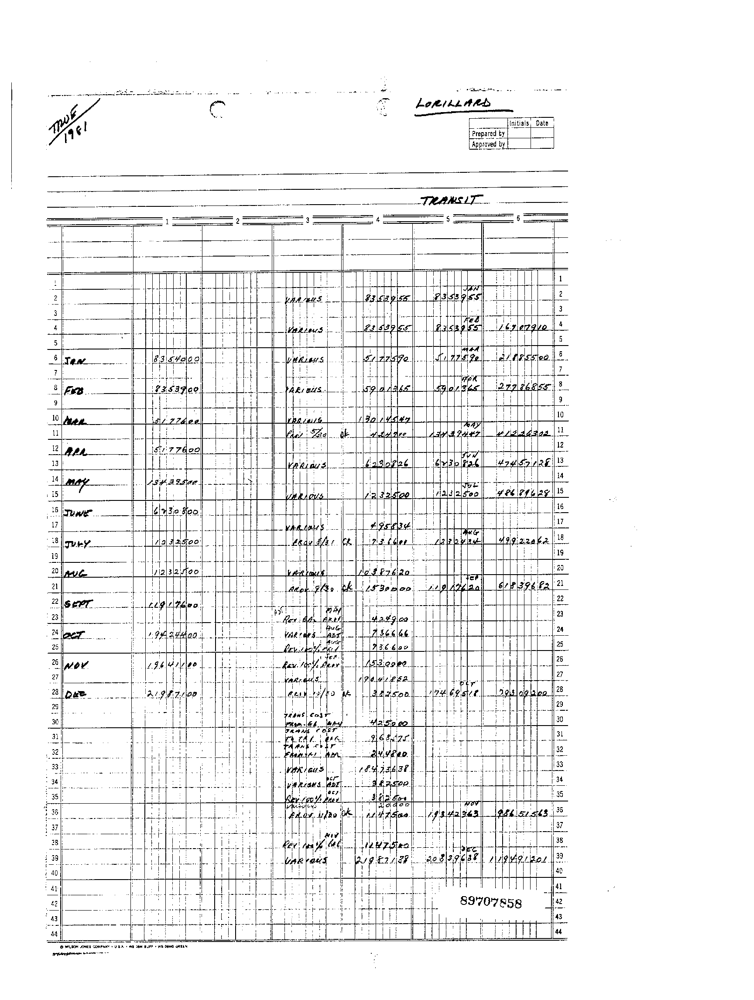}\\ GT: \hspace{0.5em}budget\\\hspace{-0.90em}Pred: \textcolor{darkred}{form}\captionsetup{font=scriptsize}\caption{}\end{subfigure} & \begin{minipage}{100px}\vspace{-10px}\textls[-25]{The image depicts a document that appears to be a spreadsheet or table, with rows and columns, and handwritten notes and numbers. The top of the page has a header with the words "Lorillard" and "Transit", and the page is filled with rows of numbers and handwritten notes. The document appears to be a financial or accounting document, possibly a ledger or a budget.}\end{minipage} &
      \begin{subfigure}[h]{68px}\centering\includegraphics[width=68px]{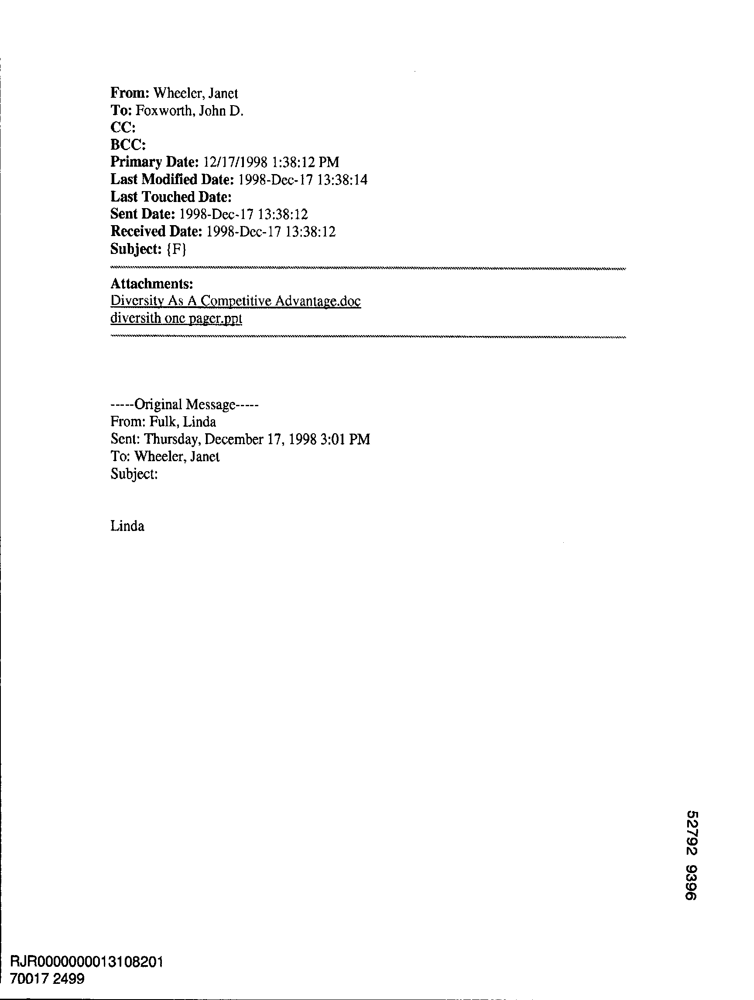}\\ GT: \hspace{0.5em}letter\\Pred: \textcolor{darkred}{email}\captionsetup{font=scriptsize}\caption{}\end{subfigure} & \begin{minipage}{100px}\vspace{-15px}\textls[-25]{The image depicts a document with a formal tone, featuring a header with the sender's and recipient's names, a subject line, and a brief summary of the content. The document is addressed to a specific individual and includes a timestamp and a CC field, which are common elements found in email messages. The presence of these features suggests that this document is an email.}\end{minipage} \\
    \end{tabular}
  \end{center}
  \caption{
    \textbf{Reasoning Examples.}
    Predictions of LLama-3.2-11B-Vision-Instruct trained with reinforcement learning (RL) with the reasoning the
    model provided in the response before giving its final answer.
  }\label{fig:results-reasoning-examples}
\end{figure}

A few examples of the reasoning given by the model are depicted in \autoref{fig:results-reasoning-examples}. The reasons
often describe what attributes were found to be relevant for the prediction, giving the user a better understanding as
to why a class was potentially chosen. For example, the news article (b) is identified by the very common layout of news
papers. Sometimes the provided reasoning makes perfect sense, but the model gets the classification wrong. This can be
seen in (c) where the reasoning ends with \textit{``[...] possibly a ledger or a budget''} but the model then predicts
\textit{form} instead of the actual class \textit{budget}. It was able to identify it correctly, but when it came to
making the final decision, it most likely put more emphasis on the other aspects that could equally be attributed to
a form. Lastly, there are some cases like (d), where not only the reasoning makes perfect sense, but also the chosen
class, yet the classification is incorrect. These stem from annotation errors, as the RVL-CDIP is known to have
inaccuracies in the annotations, particularly since there are overlapping classes that led to ambiguities, which also
made it harder for the annotators to choose the correct class.

\section{Discussions}

The primary discussion points are related to the question of whether RL is worth considering for more traditional tasks
such as classification. Based on the results presented in this paper, we think that there is merit to at least try RL
for downstream tasks that have a verifiable answer, as it seems to improve the generalisation capabilities of the model
to previously unseen data with more flexibility. There are however a few points that need to be taken into account,
which are not purely result oriented.

\subsection{Training Instability of RL}\label{sec:rl-training-instability}

RL relies heavily on the base model being able to produce at least one correct answer so that not all rewards are zero.
Thankfully, this was the case for our training, but zero advantages, which arrive when there is no variation in quality
of responses, good or bad, may occur at any point during the training, which makes the model stagnate. Without any
intervention, this could make the training unstable or even impossible.

An instability that we encountered during the training of the model on the 1\,600 images of the RVL-CDIP dataset, was
the odd behaviour of the model producing the answer before the response. This goes against the idea of increasing
test-time compute, as the final answer was already given and everything that follows does not help improve the answer
itself. Even after forcing the order with a strict format reward, it still occurred and seems to have had a negative
effect on the model, to the point where it even produced a reasoning outside of the prescribed tags and sometimes
copying the same reasoning after the answer to have it the reasoning tag. This did not occur when training on the subset
with only 10 classes, which is probably why the results in \autoref{tab:results-unseen-classes} are better for the model
that was not trained on all classes.

Since RLVR has only recently been in the spotlight for LLMs, this is expected to be improved with future research, and
hopefully methods will be developed to alleviate the instabilities in order to get the full potential out of it.

\subsection{Efficiency}

Another negative aspect of RL compared to SFT, is the fact that the training is much less efficient, since for each
batch  a group of samples needs to be generated at every iteration. The generation is usually much slower because
of the autoregressive nature of LLMs, where one token is generated at a time, resulting in many more forward passes than
during SFT. To make matters worse, the responses are generally much longer, since it also needs to generate the
reasoning. The combination of all that makes the training much less efficient. Similarly, during inference, the
responses are also longer due to the reasoning.

In terms of memory requirements, due to the use of LoRA and having access to the base model, there is no additional
memory required for any other model. The only difference is the memory required for the batches. Since the length of the
reasoning has no effective limit, unless a length reward is integrated, the batches might get much larger, which in turn
require more memory, particularly for the attention layers. This means that the batch size needs to be reduced compared
to SFT. And since the generated samples do not have a uniform length, some batches might leave a lot of head space,
which affects the hardware utilisation efficiency.

\subsection{Explainability}

The goal of having the model include the reasoning in the answer is primarily for the increased test-time compute, but
it also offers the benefit of having better explainability in order to understand why a certain decision was made.
Although we have shown some examples of the reasoning, which seem to be coherent and helpful, in particular in the cases
where the prediction was incorrect, they have not been scrutinised and evaluated in detail. As most RLVR methods do not
impose any particular demands or restrictions on the reasoning but only the format, the provided reasoning is never
guided and may not be as helpful for other tasks. This is a first step in the direction of explainability that can be
included into the training without needing an annotated reasoning dataset, but additional research is need in this area
to get the most out of it.

\section{Conclusion}

With the experiments conducted in this paper we showed that RL tends to have a better generalisation capability in the
context of document image classification, where the model adapts more easily to out-of-distribution images that come
from an entirely different era, as well as previously unseen classes. This indicates that RL is more akin to learning
the underlying fundamentals of the document classification task rather than being overly focused on the specific classes
at hand. However, it is not the case for different modalities, i.e.\ going from images to text as input or vice-versa,
where the model trained with RL struggles much more, which may be explained by the encountered training instabilities.
On the other hand, SFT is much simpler to train and generally performs strongly on in-distribution data, while suffering
more on out-of-distribution data. The biggest drawback of SFT is the memorisation aspect, which worsens its instruction
following ability as it tends to be referring back to the classes it was trained on even if they were no longer a
viable option given by the prompt.

Based on these findings, RL will hopefully be taken into consideration as a viable option for more downstream tasks,
with the understanding that the training is more challenging, which will take some tinkering to mitigate the possible
training instabilities that may arise. Additional research is necessary to arrive to a more consistent and stable
training to make the effort of RL worthwhile for a wide range of applications. Besides the improved generalisation
capabilities, a compelling aspect that emerges from RL is the reasoning that the model provides, which not only improves
its result through increased test-time compute, but also allows to get an insight into the choices that were made. 

\bibliographystyle{splncs04}
\bibliography{references}

\end{document}